\documentclass[letterpaper, 10 pt, conference]{ieeeconf}
\IEEEoverridecommandlockouts
\usepackage{xspace}
\usepackage[dvipsnames]{xcolor}
\usepackage{graphicx}
\usepackage{amsmath}
\usepackage{amssymb}
\usepackage{mathtools}
\usepackage{bm}
\usepackage{nicefrac}

\usepackage{adjustbox}
\usepackage{booktabs}
\usepackage{multirow}
\usepackage{graphicx}
\usepackage{tikz}
\usepackage{siunitx}
\usepackage{colortbl}
\usepackage{csquotes}
\usepackage{pifont}
\usepackage{xspace}
\usepackage{lipsum}
\usepackage{censor}
\usepackage[hang,flushmargin]{footmisc}

\let\labelindent\relax  
\usepackage{enumitem}

\usepackage{microtype}

\usepackage{placeins}

\usepackage[precision=2, unit=mm]{lengthconvert}

\usetikzlibrary{positioning,arrows.meta,calc,fit,backgrounds}
\usetikzlibrary{decorations.pathreplacing,calligraphy}
\usetikzlibrary{shapes.geometric}
\usetikzlibrary{patterns}

\makeatletter
\let\NAT@parse\undefined
\makeatother

\usepackage[hidelinks]{hyperref}
\usepackage{cite}
\usepackage[capitalise]{cleveref}
\crefname{section}{Sec.}{Secs.}
\Crefname{section}{Section}{Sections}
\Crefname{table}{Table}{Tables}
\crefname{table}{Tab.}{Tabs.}

\newcommand{\STQs}{STQ$_\mathrm{1}$\xspace}
\newcommand{\AQs}{AQ$_\mathrm{1}$\xspace}

\DeclareSIUnit{\pp}{\textup{p.p.}}

\let\tensor\bm
\DeclareMathSymbol{\shortminus}{\mathbin}{AMSa}{"39}

\makeatletter
\newcommand*{\@rowstyle}{}
\newcommand*{\rowstyle}[1]{%
  \gdef\@rowstyle{#1}%
  \@rowstyle\ignorespaces%
}
\newcolumntype{=}{>{\gdef\@rowstyle{}}}
\newcolumntype{+}{>{\@rowstyle}}
\makeatother

\newcommand{\tabbf}[1]{\textbf{\tablenum{#1}}}
\newcommand{\tabul}[1]{\underline{\tablenum{#1}}}

\newcommand{\cmark}{\ding{51}}%
\newcommand{\xmark}{\ding{55}}%

\newcommand{\tinycolorbox}[2]{\tikz[baseline=(a.base),inner sep=0pt]\node[fill=#1](a){#2\strut};}

\definecolor{tabhl}{HTML}{E3EEF8}%

\def\ourmethod{SGE}
\title{\LARGE\bf%
    Streaming Gaussian Encoding for 4D Panoptic Occupancy Tracking
}

\author{
    Maximilian Luz$^{1}$,
    Thomas Nürnberg$^{2}$,
    Yakov Miron$^{2,3}$,
    Abhinav Valada$^{1}$%
\thanks{
This work was funded by the Bosch Research collaboration on AI-driven automated driving.
Abhinav Valada was funded by the Deutsche Forschungsgemeinschaft (DFG, German Research Foundation) under grant number 539134284, through EFRE (FEIH\_2698644), and the state of Baden-Württemberg.
}
\thanks{$^{1}$Department of Computer Science, University of Freiburg, Germany. } 
\thanks{$^{2}$Bosch Research, Robert Bosch GmbH, Germany.}%
\thanks{$^{3}$University of Haifa, Israel}%
}

\begin{document}
\maketitle
\thispagestyle{empty}
\pagestyle{empty}


\begin{abstract}
Camera-based 4D panoptic occupancy tracking (4D-POT) is a promising paradigm for holistic scene understanding from multi-view imagery, enabling joint reasoning about geometry, semantics, and object identities across time.
Recent mask-based pipelines achieve strong performance by propagating instance queries across frames.
However, their underlying volumetric representations are typically recomputed at each timestep, limiting geometric temporal consistency, particularly under occlusion and for static scene elements.
To address this limitation, we propose a streaming Gaussian encoder that maintains a persistent volumetric scene representation for 4D-POT.
Our method models the scene as a fixed-size set of latent Gaussian queries that are propagated via ego-motion compensation and refreshed under a confidence-guided budget constraint.
Crucially, we shape Gaussian opacities through depth-based supervision to serve as proxy for visibility, enabling confidence to accumulate as a temporally aggregated measure of persistent scene support.
Together with a warmup-based multi-frame training strategy, this yields representation-level temporal coherence beyond decoder-only tracking.
Extensive experiments on Occ3D-extended nuScenes and Waymo establish a new state-of-the-art for camera-based 4D-POT, improving tracking consistency with negligible computational overhead while remaining fully compatible with existing mask-based pipelines.
We provide code and models at \url{https://sge.cs.uni-freiburg.de}.
\end{abstract}

\section{Introduction}
\label{sec:intro}

Holistic 4D scene understanding from cameras remains a key challenge for autonomous systems.
Camera-based 4D panoptic occupancy tracking (4D-POT) has recently emerged as a unified formulation that jointly models geometry, semantics, and object identities over time.
By jointly reasoning over them, recent approaches enable rich spatiotemporal awareness from multi-view imagery alone, bringing camera-only perception closer to persistent 3D scene understanding~\cite{mohan2026up,lang2024point,abdelsamad2025multi}.
Such unified 4D scene representations are particularly appealing for navigation and motion planning in dynamic environments, where both static structure and moving objects must be modeled reliably.
However, maintaining temporally consistent volumetric representations in these settings remains challenging, especially under partial observations and occlusions.\looseness=-1

Most recent 4D-POT approaches~\cite{chen2025trackocc,luz2026lags} adopt mask-based pipelines that combine a dense 3D feature volume with a set of decoder queries for semantic and instance prediction.
In these formulations, the volumetric feature grid primarily encodes geometric scene structure, while decoder queries produce embeddings that are matched against the feature volume to predict voxel masks.
For \emph{thing} classes, instance queries are propagated across frames to preserve object identity, whereas \emph{stuff} semantics are predicted from per-frame queries without temporal persistence.
More importantly, the underlying 3D feature volume, which carries the bulk of geometric information, is recomputed at each timestep.
As a result, temporal information is not explicitly preserved in the volumetric representation, and geometric consistency across frames is not explicitly enforced.
This limitation becomes particularly pronounced under occlusion and for static scene elements that are not directly associated with tracked object queries.\looseness=-1

\begin{figure}[t]
    \centering
    \input{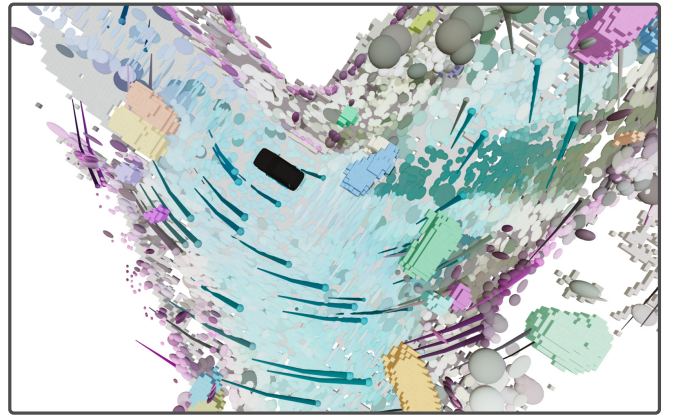}%
    \vspace{-0.5em}
    \caption{%
        Illustration of the proposed temporally coherent scene representation.
        Latent Gaussian features are visualized as PCA-colored ellipsoids.
        Darker regions indicate low opacity (a visibility proxy), revealing characteristic \enquote{shadows} behind occluding vehicles (right), whereas previously unseen regions (top left) remain clean due to confidence-based filtering.
        A random subset of Gaussian trajectories highlights temporal consistency, with trajectory width encoding confidence.
        The ego vehicle is shown in black.
    }\label{fig:teaser}%
    \vspace{-1.0em}
\end{figure}

To address these limitations, we propose a streaming Gaussian encoder that maintains a persistent and temporally consistent volumetric scene representation (illustrated in \cref{fig:teaser}), complementing decoder-level tracking with representation-level temporal coherence.
Concretely, our encoder maintains a fixed-size set of latent Gaussian queries, which are propagated across frames via ego-motion compensation and iteratively updated using the current multi-view evidence.
To keep the latent state both compact and expressive, we employ confidence-guided state management. Gaussian opacities are shaped to approximate per-frame visibility, and confidence accumulates this visibility over time, thereby retaining consistently supported structures while removing obsolete ones.
Discarded queries are replaced by newly initialized queries sampled from strong feature responses in the lifted 3D volume.
The resulting queries are jointly refined by a point-based transformer and decoded into Gaussian primitives that are splatted into the 3D feature grid.
This streaming formulation preserves geometric consistency across frames while remaining fully compatible with existing mask-based 4D-POT pipelines.

In summary, our contributions are fourfold:
\begin{itemize}[topsep=0pt]
    \item \emph{Persistent Gaussian state for 4D-POT.}
    We introduce a streaming Gaussian encoder that maintains a fixed-budget, queryable latent scene representation across timesteps, enabling representation-level temporal coherence beyond decoder-only tracking.

    \item \emph{Confidence- and visibility-guided state management.}
    We interpret Gaussian opacity as a proxy for per-frame visibility and accumulate it as evidence over time to guide pruning and refreshment of the latent state.

    \item \emph{Streaming-consistent supervision.}
    We align opacity with geometric visibility via depth-based regularization and adopt a warmup-based multi-frame training strategy to stabilize extended temporal aggregation.

    \item \emph{State-of-the-art performance with negligible overhead.}
    Our approach achieves state-of-the-art results on Occ3D nuScenes and Waymo while maintaining virtually identical inference throughput to prior mask-based pipelines.
\end{itemize}

\section{Related Work}
\label{sec:related_work}

We review prior work on camera-based occupancy prediction, temporal scene modeling, and Gaussian-based 3D representations for 4D-POT.

{\parskip=2pt
\noindent\textit{Camera-based 4D Occupancy Prediction}:
Camera-based 3D occupancy prediction has emerged as a powerful paradigm for holistic scene understanding.
With the availability of large-scale benchmarks such as nuScenes~\cite{caesar2020nuscenes} and Waymo~\cite{sun2020waymo}, together with dense 3D occupancy annotations from Occ3D~\cite{tian2023occ3d}, recent work has made substantial progress in reconstructing semantic scene geometry from multi-view imagery.
Methods including TPVFormer~\cite{huang2023TPVFormer}, SurroundOcc~\cite{wei2023SurroundOcc}, OccFormer~\cite{zhang2023OccFormer}, and SparseOcc~\cite{tang2024SparseOcc} demonstrate strong performance on camera-only semantic occupancy prediction.
More recent approaches extend this formulation toward panoptic and temporal reasoning, culminating in camera-based 4D panoptic occupancy tracking (4D-POT) as formalized by TrackOcc~\cite{chen2025trackocc}.}

{\parskip=2pt
\noindent\textit{Mask-based Panoptic Occupancy}:
Most recent 4D-POT approaches adopt mask-based architectures that combine dense volumetric features with query-based decoders, a paradigm first popularized in 2D segmentation by MaskFormer~\cite{cheng2021MaskFormer} and Mask2Former~\cite{cheng2022Mask2Former}.
OccFormer~\cite{zhang2023OccFormer} and COTR~\cite{ma2024COTR} bring this query-driven formulation to 3D perception, which is further developed into panoptic occupancy approaches, including SparseOcc~\cite{tang2024SparseOcc} and PaSCo~\cite{cao2024PaSCo}.
Recent 4D-POT systems, such as TrackOcc~\cite{chen2025trackocc} and LaGS~\cite{luz2026lags} build on this design, in which decoder queries produce mask embeddings that are matched against a dense 3D feature volume. 
While propagating instance queries across frames preserves object identities, the underlying volumetric features are typically recomputed at each timestep.
Some methods incorporate short-term temporal fusion by aligning and aggregating features from adjacent frames~\cite{ma2024COTR,huang2022BEVDet4D,li2022BEVFormer}. However, the resulting 3D representation remains largely frame-centric rather than persistently maintained.
Consequently, temporal consistency is largely handled at the decoder level, and geometric coherence of the volumetric representation is not explicitly enforced.
This limitation becomes particularly pronounced under occlusion and for static scene elements that are not directly associated with tracked object queries.}

{\parskip=2pt
\noindent\textit{Temporal Scene Modeling}: 
A large body of work explores temporal fusion for camera-based 3D perception.
BEV-based methods such as BEVFormer~\cite{li2022BEVFormer}, BEVDet4D~\cite{huang2022BEVDet4D}, and SOLOFusion~\cite{park2022SoloFusion} aggregate multi-frame image features to improve bird’s-eye-view representations.
Subsequent approaches including StreamPETR~\cite{wang2023StreamPETR}, Sparse4D~\cite{lin2022Sparse4D}, and UniOcc~\cite{wang2025UniOcc} investigate streaming-friendly and memory-enhanced multi-frame perception.
In parallel, occupancy-focused methods such as OccFlowNet~\cite{boeder2024OccFlowNet} and ForecastOcc~\cite{mohan2026ForecastOcc} incorporate additional temporal cues for dynamic scene reasoning.
Despite these advances, most approaches rely on short-term feature aggregation or recurrent feature-volume updates, yielding frame-centric representations that are recomputed at each timestep.
In contrast, our approach maintains a persistent volumetric scene state via latent Gaussian queries.}

{\parskip=2pt
\noindent\textit{Gaussian Scene Representations}: 
Gaussian-based representations have recently emerged as an efficient alternative for 3D scene modeling.
3D Gaussian Splatting~\cite{schmidt2024nerf} demonstrates high-fidelity real-time rendering using explicit Gaussian primitives, but focuses primarily on offline scene reconstruction, with extensions for motion-aware reconstruction~\cite{luiten2024Dynamic3DGaussians}.
Subsequent works such as GaussianFormer~\cite{huang2025GaussianFormer}, GaussianFormer-2~\cite{huang2025GaussianFormer2}, GaussianWorld~\cite{zuo2025GaussianWorld}, and Chorus~\cite{li2025Chorus} adapt Gaussian representations for online scene understanding and occupancy prediction.
Most closely related to our work, LaGS~\cite{luz2026lags} introduces latent Gaussian queries within a per-frame encoder and employs mask-based decoding for 4D-POT.
However, temporal consistency is primarily handled at the decoder level, leaving the encoded scene state itself temporally independent across frames.
In contrast, our approach maintains temporal coherence directly within the volumetric representation, enabling representation-level temporal consistency.
}

Overall, existing methods predominantly rely on frame-centric volumetric representations, with temporal reasoning largely confined to the decoder, where geometric scene structure is only indirectly represented.
Persistent scene representations for streaming 4D-POT remain largely unexplored.

\section{Method}
\label{sec:method}

\begin{figure*}[t]
    \centering
    \input{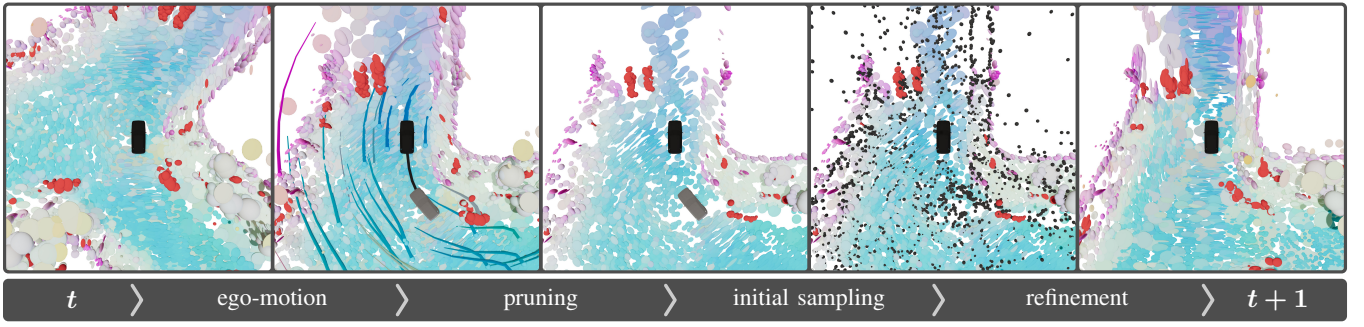}%
    \caption{
        Illustration of the proposed streaming Gaussian encoder across timesteps.
        Starting from the PCA-colored Gaussian state at time $t$ (thing-class Gaussians shown in red), queries are propagated via ego-motion compensation (EMC), highlighted by the overlaid trajectories.
        Confidence- and visibility-guided pruning then removes unreliable Gaussians, particularly in previously unseen regions shadowed by dynamic objects (top left), while retaining well-supported structure in observed areas (bottom right).
        New queries are initialized at high-response feature locations (black dots) and jointly refined, yielding the updated Gaussian state at time $t{+}1$.
        The current ego vehicle is shown in black, and the previous pose is shown in gray.
    }%
    \label{fig:architecture}%
    \vspace{-0.7em}
\end{figure*}

We address camera-based 4D panoptic occupancy tracking with a temporal Gaussian encoder that maintains a persistent, queryable 3D scene representation over time.
\cref{ssec:task} formalizes the task, followed by an architectural overview in \cref{ssec:overview}.
\cref{ssec:encoder} then details the proposed encoder, including the latent state representation, feature-aware ego-motion compensation, and confidence-guided state management under a fixed query budget.
Finally, \cref{ssec:training} describes the depth-regularized and streaming-aware training strategy.

\subsection{Task Definition}
\label{ssec:task}

Camera-based 4D panoptic occupancy tracking \cite{chen2025trackocc} requires joint prediction of 3D occupancy, semantics, and temporally consistent instance assignments from multi-view image observations.
In this work, the task is formulated as a streaming prediction problem.
At each timestep $t$, the input consists of synchronized multi-view RGB images $\tensor{I}_t = \{I_t^j\}_{j=1}^{N}$, where $N$ is the number of cameras, together with known camera intrinsics and extrinsics, as well as ego-motion from $t-1$ to $t$.
Approaches may optionally maintain an aggregated temporal state.
Formally, given a set of classes $\mathcal{C}$ (including free space, tracked \emph{thing}- and static/background \emph{stuff}-type categories), the task is to assign a tuple $(c_{\tensor{x}, t}, i_{\tensor{x}, t}) \in \mathcal{C} \times \mathbb{N}$ to each spatial location $\tensor{x} \in \mathcal{V}$, where $\mathcal{V} \subset \mathbb{R}^3$ denotes the spatial region of interest.
Here, $c_{\tensor{x},t}$ represents the semantic occupancy label and $i_{\tensor{x},t}$ the associated instance identity.
Notably, instance assignments are defined only on \emph{thing} classes and must remain temporally consistent across timesteps.
For \emph{stuff} classes, instance identities are not defined.
$\mathcal{V}$ is discretized as a voxel grid of size $X \times Y \times Z$.

\subsection{Overall Architecture}
\label{ssec:overview}

Our approach builds on an existing 4D panoptic occupancy pipeline \cite{luz2026lags} and introduces a novel temporal Gaussian encoder (\cref{ssec:encoder}) to learn temporally consistent volumetric representations.
Given a set of multi-view input images $\tensor{I}_t$ at timestep $t$, image features are first extracted and lifted into 3D using low-resolution depth predictions, yielding an initial feature volume.
This volume is refined into a multi-scale 3D feature pyramid, which is subsequently processed by our temporal Gaussian encoder.
The encoder aggregates information over time, maintaining a temporally consistent latent state across frames.
The resulting representation is converted into a dense 3D feature volume at the target resolution $X \times Y \times Z$, which serves as input to a mask-based 4D panoptic occupancy decoder \cite{luz2026lags}, to produce semantic occupancy and temporally consistent instance assignments.

\subsection{Temporal Gaussian Encoder}
\label{ssec:encoder}

Our temporal Gaussian encoder is designed to learn a temporally consistent, queryable 3D representation by maintaining and updating a set of latent queries over time.
These queries act as a persistent latent state (\cref{sssec:state}) and are decoded into Gaussian primitives that carry localized geometric and semantic information, yielding a continuous volumetric representation.
This formulation builds on the temporally independent Gaussian-based LaGS encoder \cite{luz2026lags}, which constructs volumetric representations from multi-view observations via query sampling, refinement, and splatting, and extends it to the temporal domain.

In the single-frame setting, LaGS initializes query features $\tensor{q}_k$ by sampling from the multi-scale 3D feature pyramid obtained after the lifting step.
The queries are serialized via space-filling curves and refined using a point-based transformer, where different scales are processed as separate streams that interact via window-based multi-stream attention.
Within each scale, self-attention models interactions between queries, while spatial image cross-attention enables efficient implicit 2D-to-3D lifting.
After refinement, each query $\tensor{q}_k$ is decoded into Gaussian parameters, including center $\tensor{\mu}_k$, covariance, opacity $\alpha_k$, and embedding feature $\tensor{e}_k$.
The embedding features $\tensor{e}_k$ are then splatted into a dense 3D feature volume using the corresponding Gaussian primitives.

Building on this formulation, our temporal encoder introduces a persistent query state and a streaming update mechanism across timesteps.
As new observations arrive, the latent queries are iteratively propagated, refreshed, and refined, enabling efficient multi-view aggregation while preserving temporal consistency.
The update proceeds in three stages, illustrated in \cref{fig:architecture}.
First, queries from the previous timestep are propagated using feature-aware ego-motion compensation to maintain spatial alignment (\cref{sssec:emc}).
Second, the latent state is refreshed under a fixed query budget via confidence-guided pruning and magnitude-based re-initialization (\cref{sssec:pruning}), allowing obsolete queries to be replaced by newly observed structures.
Finally, the resulting set of queries is jointly refined using the same point-based transformer architecture as in the single-frame encoder.
The refined queries are then stored for the next timestep, decoded into Gaussian primitives, and splatted into a 3D voxel grid for the downstream 4D-POT decoder.

\subsubsection{State Representation}
\label{sssec:state}

The encoder maintains a persistent set of $K$ latent queries $\{\tensor{q}_k\}_{k=1}^K$ and corresponding reference positions $\tensor{p}_k \in \mathbb{R}^3$, together forming the temporal state that is propagated and updated over time.
Each query can be decoded into a Gaussian primitive that provides an explicit geometric interpretation of the latent state.
Specifically, a query $\tensor{q}_k$ parameterizes a Gaussian with center $\tensor{\mu}_k$, covariance (represented via scale and rotation), opacity $\alpha_k$, embedding features $\tensor{e}_k$, and semantic predictions (class logits) $\tensor{z}_k$.
The Gaussian center is defined relative to the reference position, i.e., $\tensor{\mu}_k = \tensor{p}_k + \tensor{\delta}_k$ where $\tensor{\delta}_k$ is decoded from the Gaussian, enabling stable spatial tracking across frames.
These Gaussian attributes are used throughout the update procedure: centers define spatial locations for ego-motion compensated propagation, opacities provide a measure of importance for pruning, and embedding features contribute to the final volumetric representation.

\subsubsection{Feature-Aware Ego-Motion Compensation}
\label{sssec:emc}

Each query represents a persistent spatial feature across timesteps.
As the camera moves, ego-motion compensation is therefore required during propagation to preserve this correspondence and maintain spatial alignment of the latent state.
Given the ego-motion between frames $t-1$ and $t$ as a rigid transform $T = [R \mid \tensor{t}]$, updated query reference positions are obtained by transforming the decoded Gaussian centers:
\begin{equation}
    \tensor{p}_k^{(t)} = T \cdot \tensor{\mu}_k^{(t-1)}.
\end{equation}
To update position-dependent feature components, we employ a FiLM-based modulation scheme \cite{perez2018FiLM}.
Global modulation parameters $\tensor{\gamma}, \tensor{\beta} \in \mathbb{R}^d$ are predicted from ego-motion magnitude $m = \|\tensor{t}\|_2$ and rotation $\theta = \arccos\left(\nicefrac{(\operatorname{tr}(R) - 1)}{2}\right)$, providing a compact and sufficient summary of the global motion, as $[\boldsymbol{\gamma}; \boldsymbol{\beta}] = \operatorname{MLP}_{\text{mod}}([m;\theta])$, with $\tensor\gamma$ bias-initialized to $\tensor{1}$ and $\tensor\beta$ to $\tensor{0}$.
Local motion cues are encoded per query as 
\begin{equation}
    \tensor{m}_k = \operatorname{MLP}_{\text{mot}}\left(\tensor{p}_k^{(t)} - \tensor{\mu}_k^{(t-1)}\right)
\end{equation}
and used to compute feature updates
\begin{equation}
    \tensor{\delta}_k = \operatorname{MLP}_{\text{upd}}\left(\left[\tensor{q}_k^{(t-1)};\, \tensor{m}_k\right]\right).
\end{equation}
Finally, query features are updated via a gated, FiLM-modulated residual step:
\begin{align}
    \tensor{q}_k^{(t)} = \tensor{q}_k^{(t-1)} + \sigma(g)\cdot \left( \tensor{\gamma} \odot \tensor\delta_k + \tensor{\beta} \right),
\end{align}
where $\sigma(\cdot)$ is the sigmoid function and $g$ is a learned scalar initialized to yield small updates at early stages of training, promoting stable learning dynamics.

\subsubsection{Confidence-Guided Pruning}
\label{sssec:pruning}

While ego-motion compensation preserves spatial alignment across frames, the latent state must be refreshed under a fixed query budget as new regions become visible and old ones become obsolete.
However, pruning based solely on instantaneous observations is insufficient, as temporarily occluded yet valid structures may be removed prematurely.
To address this, we assign each query a confidence score $\rho_k$ that accumulates visibility evidence over time.
Here, Gaussian opacity $\alpha_k$ serves as a proxy for per-frame visibility, whose semantics are shaped through depth-based supervision (cf.~\cref{sssec:depth}).

Confidence is recursively updated to reflect temporal reliability.
We compute a per-frame observation score $o_k$ from Gaussian opacity $\alpha_k$, modulated by distance-based falloff $d_k$ and gated by field-of-view visibility:
\begin{equation}
    o_k = \alpha_k \cdot d_k \cdot \mathbf{1}_{\text{FoV}}(\tensor{\mu}_k), \quad
    d_k = \frac{1}{1 + \nicefrac{\|\tensor{\mu}_k\|_2}{\lambda_{\text{dist}}}},
\end{equation}
where $\lambda_{\text{dist}}$ controls the distance-based decay.
The confidence is then updated as
\begin{equation}
    \rho_k^{(t)} = \rho_k^{(t-1)} + \lambda \, o_k^{(t)} \left(1 - \rho_k^{(t-1)}\right), \quad \rho_k^{(0)} = 0,
\end{equation}
with $\lambda$ controlling accumulation speed.

Since ego-motion compensation accounts only for camera motion, Gaussians belonging to dynamic classes (e.g., vehicles or pedestrians) may become spatially inconsistent over time.
We therefore apply an additional per-frame decay:
\begin{equation}
    \rho_k^{(t)} =
    \rho_k^{(t)} \cdot
    \begin{cases}
        \lambda_{\text{dyn}}, & \text{if } \hat{c}_k \in \mathcal{C}_{\text{dyn}}, \\
        1, & \text{otherwise},
    \end{cases}
\end{equation}
where $\hat{c}_k = \arg\max \tensor{z}_k$ denotes the predicted class of the $k$-th Gaussian and $\lambda_{\text{dyn}}$ controls dynamic decay strength.

{\parskip=2pt
\noindent\textit{Budget-Constrained State Refreshment}:
Given the confidence estimates $\rho_k$, the latent state is refreshed under a fixed budget of $K$ queries.
Queries are first filtered using temporal confidence, instantaneous opacity, and spatial relevance:
\begin{equation}
    \mathcal{K}' = \left\{\; k \;\middle|\; \left( \rho_k^{(t)} \ge \tau_{\rho} \;\lor\; \alpha_k^{(t)} \ge \tau_\alpha \right) \;\land\; \tensor{p}_k \in \mathcal{V}_p \;\right\}.
\end{equation}
If $|\mathcal{K}'| > K - M$, we enforce the query budget via hybrid rank-based sampling. Otherwise, all eligible queries are retained. 
To favor queries supported by either persistent confidence or strong instantaneous evidence, we define the hybrid rank
\begin{equation}
    r_k = \max\!\left\{
        \operatorname{rank}\left(\rho_k\right)\!,
        \,\operatorname{rank}\left(\alpha_k\right)
        \right\}.
\end{equation}
The retained set of $K - M$ queries is then sampled as
\begin{equation}
    \mathcal{K}_{p} \sim \mathcal{M}(\tensor{\pi},\, K - M),
    \quad
    \pi_k = \nicefrac{r_k}{\sum_j r_j},
\end{equation}
where $\mathcal{M}(\cdot)$ denotes multinomial sampling without replacement, ensuring that at least $M$ slots remain available for new queries in the current frame.
In practice, we find that $\tau_\alpha{=}0.01$ by itself is a reliable discriminator for visibility.
Consequently, combining this with $\tau_\rho{=}0.01$, $\lambda{=}0.8$, and $\lambda_{\text{dist}}{=}20$ retains Gaussians that are either visible in the current frame or have been observed well in the past.
Additionally, we set $\lambda_{\text{dyn}}{=}0.9$.
}

{\parskip=2pt
\noindent\textit{Initialization of New Queries}:
To replenish the pruned slots, new queries are initialized at locations with strong responses in the lifted 3D feature volume.
Given per-voxel feature vectors $\tensor{f}_k$, we define sampling priorities based on feature magnitude and sample indices via multinomial sampling:
\begin{equation}
    \mathcal{K}_{n} \sim \mathcal{M}(\tensor{\pi},\, K - |\mathcal{K}_p|),
    \quad
    \pi_k = \nicefrac{\|\tensor{f}_k\|_2}{\sum_j \|\tensor{f}_j\|_2}.
\end{equation}
The final query set is given by $\mathcal{K} = \mathcal{K}_{p} \cup \mathcal{K}_{n}$,
ensuring a fixed budget of $|\mathcal{K}| = K$ queries at each timestep.
}

\subsection{Training and Supervision}
\label{ssec:training}

Our temporal encoder is trained end-to-end as part of the overall model.
To supervise the persistent latent state, we combine geometric supervision with a streaming-aware training strategy.
Specifically, we employ (i) a depth splatting loss that regularizes the lifted 3D representation and shapes Gaussian opacities, and (ii) a multi-frame training procedure that exposes the encoder to realistic streaming inputs while keeping memory overhead low.

\subsubsection{Depth Splatting Supervision}
\label{sssec:depth}

Since confidence aggregates observation quality over time, confidence-guided pruning requires opacity to reflect per-frame visibility.
We therefore align opacity with geometric support by regularizing Gaussian opacities using sparse LiDAR depth.
To this end, decoded Gaussians are splatted onto the image plane, encouraging surface-consistent Gaussians to attain high opacity while suppressing those inconsistent with measured geometry or located in occluded regions.
Consequently, opacity serves as a proxy for per-frame visibility during confidence updates (cf.~the characteristic shadows in \cref{fig:teaser}).

Depth is rendered by alpha compositing the decoded Gaussians onto the image plane and supervised using
\begin{equation}
    \mathcal{L}_{\text{depth}}
    =
    \frac{1}{|\Omega|}
        \sum_{\tensor{u} \in \Omega}
        \left| \hat{D}(\tensor{u}) - D(\tensor{u}) \right|,
\end{equation}
where $\hat{D}(\tensor{u})$ denotes the rendered depth at pixel $\tensor{u}$, $D(\tensor{u})$ the corresponding sparse LiDAR depth, and $\Omega$ the set of pixels with valid LiDAR measurements.

\subsubsection{Multi-Frame Training}
\label{sssec:multiframe}

We follow a two-stage training protocol of LaGS, which consists of (i) single-frame pre-training for detection and (ii) multi-frame training for instance tracking.
During the tracking stage, gradients are detached across consecutive frames to bound memory usage, while supervision is provided by the panoptic decoder as well as Gaussian-based auxiliary losses~\cite{luz2026lags}.
However, this training scheme does not explicitly initialize the temporal state of the proposed encoder.
To expose the model to a well-formed temporal context, we introduce \emph{warmup frames} in both training stages.
Specifically, a small number of preceding frames are processed by the temporal encoder to initialize its latent state.
These frames are excluded from loss computation and are not passed to the decoder.
Gradients are detached during warmup steps, ensuring they serve only to initialize temporal context without increasing memory consumption.
This strategy enables training under realistic streaming conditions, and we find that, despite the absence of cross-frame gradient flow, it is sufficient for the encoder to learn temporally consistent update dynamics.
\section{Experiments}
\label{sec:experiments}

We evaluate our proposed streaming Gaussian encoder on camera-based 4D panoptic occupancy tracking using the Occ3D-extended nuScenes and Waymo benchmarks.
\cref{ssec:datasets,ssec:implementation} describe datasets, metrics, and implementation details.
Main quantitative results are presented in \cref{ssec:results}, followed by comparisons to single-frame occupancy methods (\cref{ssec:results_occ3d}), qualitative analyses (\cref{ssec:qualitative}), and ablations (\cref{ssec:ablations}) assessing contributions of each component.

\subsection{Datasets and Evaluation Metrics}
\label{ssec:datasets}

We evaluate on the nuScenes~\cite{caesar2020nuscenes} and Waymo~\cite{sun2020waymo} benchmarks using occupancy annotations from Occ3D~\cite{tian2023occ3d}.
We follow the standard camera-based 4D panoptic occupancy tracking protocol established by TrackOcc~\cite{chen2025trackocc}.
For both datasets, the evaluation volume spans \SI{-40}{\meter} to \SI{40}{\meter} in the horizontal plane and \SI{-1}{\meter} to \SI{5.4}{\meter} vertically, discretized with a voxel size of \SI{0.4}{\meter}.
Performance is primarily measured using Segmentation and Tracking Quality (STQ)~\cite{weber2021step,chen2025trackocc}, defined as the geometric mean of Segmentation Quality (SQ) and Association Quality (AQ), jointly evaluating semantic occupancy and temporal instance consistency.
To assess single-frame panoptic performance without temporal association, we additionally report {\STQs} and {\AQs}, the non-temporal counterparts of STQ and AQ \cite{luz2026lags}.
We report the binary IoU for free/occupied occupancy prediction.

\subsection{Implementation Details}
\label{ssec:implementation}

Following LaGS~\cite{luz2026lags}, we use an ImageNet-pretrained VoVNetV2-99 (V99) backbone~\cite{lee2019VoVNetV2} with input resolutions of $800{\times}320$ for nuScenes and $704{\times}256$ for Waymo, and adopt the same multi-view lifting, 3D feature construction, architecture, optimizer, learning schedule, and loss weights.
Unless stated otherwise, all experiments use the V99 backbone; additional results with a ResNet-50 (R50) backbone are reported for comparison.
Based on empirical analysis, we employ our encoder only at the fine stream of the two-stream LaGS variant (LaGS-2s), as the coarse stream fails to produce meaningful Gaussian primitives.
We maintain a fixed query budget of $K{=}8192$, with at least $M{=}2048$ newly initialized queries per frame and latent dimension $d{=}256$.

We train the models on $8$ NVIDIA L40 GPUs for $24$ epochs (12 detection pre-training and 12 tracking epochs), with a batch size of $1$ per GPU.
Each training sequence contains $8$ consecutive frames.
During detection pre-training, sequences comprise $7$ warmup frames followed by one supervised frame, while tracking training uses $5$ warmup and $3$ supervised frames.
Gradients are detached between frames. 
The Gaussian depth loss is applied with weight $1.0$ at $\nicefrac{1}{4}$ input resolution.

\subsection{Main Results}
\label{ssec:results}

Results for 4D panoptic occupancy tracking on Occ3D-nuScenes and Occ3D-Waymo are presented in \Cref{tab:results_nusc,tab:results_waymo}.
Our method outperforms prior work across all major temporal (STQ, AQ) and single-frame (\STQs, \AQs) metrics.
On Occ3D-nuScenes, our approach improves STQ from \num{32.3} to \num{34.4} over LaGS-2s, with corresponding gains in AQ, semantic quality, and binary occupancy IoU.
Improvements are observed for both \emph{thing} and \emph{stuff} classes, indicating that the proposed streaming representation benefits both dynamic object reasoning and static scene modeling.
On Occ3D-Waymo, our method similarly advances the state of the art, achieving \num{21.9} STQ (\num[explicit-sign=+]{0.7} over LaGS-2s) together with improvements in AQ, \STQs, \AQs, and mIoU.
Although the gains are smaller than on nuScenes, they are observed across both temporal and semantic metrics.
We attribute the reduced margin to Waymo's predominantly forward-driving scenarios and lower occlusion frequency, which diminish the benefits of extended temporal aggregation.

\begin{table}[!tb]
    \centering
    \caption{4D-POT performance on Occ3D-nuScenes.}
    \label{tab:results_nusc}%
    \vspace{-0.75em}
    \footnotesize
\setlength{\tabcolsep}{2.5pt}  
\begin{tabular}{=+l*{8}{+S[table-format=2.1,detect-all]}}
\toprule
    & & & & & \multicolumn{3}{c}{mIoU} & \\
\cmidrule(lr){6-8}
    \textbf{Approach} & {STQ} & {AQ} &  {\STQs} & {\AQs} & {All} & {Things} & {Stuff} & {IoU} \\
\midrule

Per-Frame \cite{luz2026lags} &
     9.0 &  2.5 & 21.8 & 14.7 & 32.5 & 26.4 & 41.2 & 63.2
    \\

MinVIS$^\dagger$ \cite{huang2022minvis} &
    11.8 &  4.3 & 21.8 & 14.7 & 32.5 & 26.4 & 41.2 & 63.2
    \\

CTVIS$^\dagger$ \cite{ying2023ctvis} &
    11.4 &  3.9 & 22.5 & 15.4 & 33.0 & 27.0 & 41.5 & 63.8
    \\

4D-LCA$^\dagger$ \cite{aygun20214dpls} &
    12.5 &  4.8 & 21.8 & 14.7 & 32.5 & 26.4 & 41.2 & 63.2
    \\

AB3DMOT$^\dagger$ \cite{weng2020ab3dmot} &
    13.1 &  5.3 & 21.8 & 14.7 & 32.5 & 26.4 & 41.2 & 63.2
    \\

TrackOcc$^\dagger$ \cite{chen2025trackocc} &
    12.2 &  4.7 & 19.7 & 12.1 & 32.1 & 25.3 & 41.8 & 63.7
    \\
    
LaGS-2s R50 \cite{luz2026lags}
    & \tabul{27.4}
    & \tabul{21.4}
    & \tabul{31.4}
    & \tabul{27.2}
    & \tabul{36.2}
    & \tabul{31.4}
    & \tabul{43.0}
    & \tabul{64.4}
    \\

\rowcolor{tabhl}
{\ourmethod} R50 (Ours)
    & \tabbf{31.8}
    & \tabbf{24.8}
    & \tabbf{36.0}
    & \tabbf{31.8}
    & \tabbf{40.9}
    & \tabbf{36.9}
    & \tabbf{46.5}
    & \tabbf{69.5}
    \\

\arrayrulecolor{lightgray}
\midrule
\arrayrulecolor{black}

TrackOcc V99$^{\dagger}$ \cite{chen2025trackocc}
    & 15.5
    &  6.5
    & 23.8
    & 15.3
    & 37.0
    & 31.0
    & 45.6
    & 67.0
    \\

LaGS-2s V99 \cite{luz2026lags}
    & \tabul{32.3}
    & \tabul{25.6}
    & \tabul{36.4}
    & \tabul{32.5}
    & \tabul{40.7}
    & \tabul{37.0}
    & \tabul{46.0}
    & \tabul{67.3}
    \\

\rowcolor{tabhl}
{\ourmethod} V99 (Ours)
    & \tabbf{34.4}
    & \tabbf{27.2}
    & \tabbf{38.5}
    & \tabbf{34.1}
    & \tabbf{43.5}
    & \tabbf{39.5}
    & \tabbf{49.1}
    & \tabbf{71.7}
    \\

\bottomrule
\end{tabular}%

\vspace{0.5em}
\begin{minipage}[t]{0.99\linewidth}
    \scriptsize
    $\dagger$: As reported by \cite{luz2026lags}.%
\end{minipage}%
\end{table}

\begin{table}[!tb]
    \centering
    \caption{4D-POT performance on Occ3D-Waymo.}%
    \label{tab:results_waymo}%
    \vspace{-0.75em}
    \footnotesize
\setlength{\tabcolsep}{2.5pt}  
\begin{tabular}{=+l*{8}{+S[table-format=2.1,detect-all]}}
\toprule
    & & & & & \multicolumn{3}{c}{mIoU} & \\
\cmidrule(lr){6-8}
    \textbf{Approach} & {STQ} & {AQ} &  {\STQs} & {\AQs} & {All} & {Things} & {Stuff} & {IoU} \\
\midrule

Per-Frame \cite{luz2026lags} &
     9.1 &  4.0 & 18.1 & 15.6 & 20.9 & 21.9 & 20.6 & 58.4
    \\

MinVIS$^\dagger$ \cite{huang2022minvis} &
    11.0 &  5.8 & 18.1 & 15.6 & 20.9 & 21.9 & 20.6 & 58.4
    \\

CTVIS$^\dagger$ \cite{ying2023ctvis} &
    12.5 &  7.3 & 18.7 & 16.5 & 21.2 & 22.3 & 20.9 & 59.6
    \\

4D-LCA$^\dagger$ \cite{aygun20214dpls} &
    12.1 &  7.0 & 18.1 & 15.6 & 20.9 & 21.9 & 20.6 & 58.4
    \\

AB3DMOT$^\dagger$ \cite{weng2020ab3dmot} &
    13.3 &  8.5 & 18.1 & 15.6 & 20.9 & 21.9 & 20.6 & 58.4
    \\

TrackOcc$^\dagger$ \cite{chen2025trackocc} &
    15.2 & 10.7 & 18.1 & 15.2 & 21.6 & 21.9 & \tabul{21.5} & \tabul{60.1}
    \\

LaGS-2s R50 \cite{luz2026lags}
    & \tabul{18.4}
    & \tabul{15.3}
    & \tabul{20.6}
    & \tabul{19.2}
    & \tabul{22.0}
    & \tabul{24.3}
    & 21.4
    & 59.8
    \\
    
\rowcolor{tabhl}
{\ourmethod} R50 (Ours)
    & \tabbf{20.0}
    & \tabbf{16.8}
    & \tabbf{22.1}
    & \tabbf{20.7}
    & \tabbf{23.7}
    & \tabbf{25.8}
    & \tabbf{23.2}
    & \tabbf{60.3}
    \\

\arrayrulecolor{lightgray}
\midrule
\arrayrulecolor{black}

TrackOcc V99$^\dagger$ \cite{chen2025trackocc}
    & 16.7
    & 11.7
    & 20.0
    & 16.7
    & 23.9
    & 24.7
    & \tabul{23.7}
    & \tabbf{62.9}
    \\

LaGS-2s V99 \cite{luz2026lags}
    & \tabul{21.2}
    & \tabul{18.3}
    & \tabul{23.6}
    & \tabul{22.6}
    & \tabul{24.6}
    & \tabbf{28.1}
    & \tabul{23.7}
    & 61.5
    \\
    
\rowcolor{tabhl}
{\ourmethod} V99 (Ours)
    & \tabbf{21.9}
    & \tabbf{19.0}
    & \tabbf{24.1}
    & \tabbf{23.0}
    & \tabbf{25.3}
    & \tabul{27.8}
    & \tabbf{24.6}
    & \tabul{61.9}
    \\

\bottomrule
\end{tabular}%

\vspace{0.5em}
\begin{minipage}[t]{0.99\linewidth}
    \scriptsize
    $\dagger$: As reported by \cite{luz2026lags}.%
\end{minipage}%
\vspace{-2.0em}
\end{table}

Despite introducing temporal state management, our encoder incurs negligible computational overhead.
On Occ3D-nuScenes, our method processes \num{2.94} samples/s at inference on an NVIDIA L40 GPU, compared to \num{3.00} for LaGS, corresponding to a marginal slowdown of approximately \SI{2}{\percent}.
This demonstrates that persistent volumetric modeling can be achieved with virtually no throughput penalty while substantially improving temporal consistency.

\subsection{Comparison to Single-Frame 3D Occupancy Methods}
\label{ssec:results_occ3d}

\begin{table}[t]
    \centering
    \caption{3D occupancy prediction performance on Occ3D-nuScenes.}
    \vspace{-0.75em}
    \footnotesize
\setlength{\tabcolsep}{9pt}  
\begin{tabular}{=+lc*{2}{+S[table-format=2.1,detect-all]}}

\toprule
\textbf{Approach} & Image Backbone
    & {mIoU}
    & {IoU} \\
\midrule

TPVFormer \cite{huang2023TPVFormer}             & ResNet-50  &        34.2  &        66.8  \\
SurroundOcc \cite{wei2023SurroundOcc}           & ResNet-101 &        34.6  &        65.5  \\
OccFormer \cite{zhang2023OccFormer}             & ResNet-50  &        37.4  &        70.1  \\
BEVDet4D \cite{huang2022BEVDet4D}               & ResNet-50  &        39.3  &        73.8  \\
BEVDet4D + COTR \cite{ma2024COTR}               & ResNet-50  & \tabul{44.5} & \tabbf{75.0} \\
BEVDet4D + COTR \cite{ma2024COTR}               & Swin-B     & \tabbf{46.2} & \tabul{74.9} \\

\rowcolor{gray!15}
BEVDet4D + COTR$^{\ddagger}$ \cite{ma2024COTR}  & ResNet-50  &        40.2  &         71.7 \\

\midrule

TrackOcc$^{\dagger}$ \cite{chen2025trackocc}    & ResNet-50   &        32.1  &        63.7  \\
LaGS-2s \cite{luz2026lags}                      & VoVNetV2-99 & \tabul{40.7} & \tabul{67.3} \\

\rowcolor{tabhl}
{\ourmethod} (Ours)                             & VoVNetV2-99 & \tabbf{43.5} & \tabbf{71.7} \\

\bottomrule
\end{tabular}%

\vspace{0.5em}
\begin{minipage}[t]{0.99\linewidth}
    \scriptsize
    Top: 3D semantic occupancy prediction methods without instance segmentation or tracking.
    Bottom: 4D panoptic occupancy tracking methods requiring instance-level modeling and temporal association.
    $\dagger$: As reported by \cite{luz2026lags}.
    Gray row/$\ddagger$: Restricted baseline without CFSG, representing the closest comparable setting.

\end{minipage}%
\vspace{-1.1em}
    \label{tab:results_occ3d}
\end{table}

While our primary focus is camera-based 4D panoptic occupancy tracking, we additionally compare against semantic occupancy methods on Occ3D (\cref{tab:results_occ3d}).
Because 4D-POT jointly predicts geometry, semantics, and temporal identities, it constitutes a strictly more challenging task than semantic occupancy alone.
Nevertheless, our method achieves \num{43.5} mIoU and \num{71.7} IoU, substantially outperforming prior 4D-POT approaches such as TrackOcc and LaGS-2s.
Despite optimizing for the more demanding panoptic and temporal setting, our model remains highly competitive with dedicated semantic occupancy approaches, even surpassing BEVDet4D+COTR without coarse-to-fine semantic grouping (CFSG)~\cite{ma2024COTR}, the closest purely semantic counterpart.
These results significantly narrow the gap between 3D and 4D occupancy approaches.

\subsection{Qualitative Results}
\label{ssec:qualitative}

\begin{figure}[tb]
    \centering
    \input{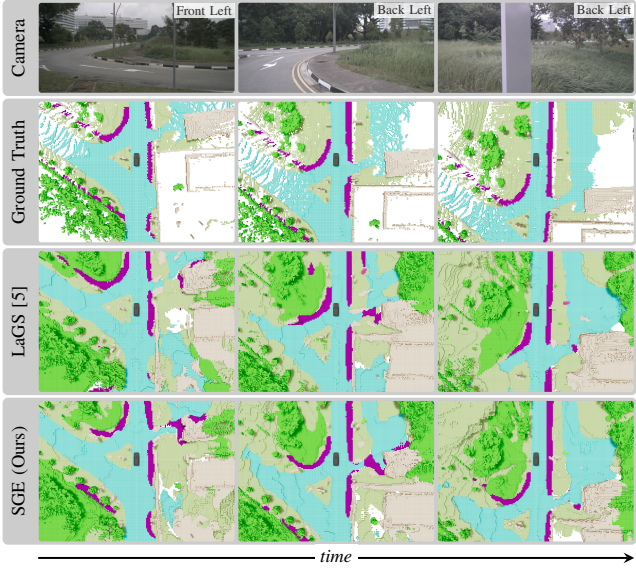}%
    \vspace{-0.75em}
    \caption{%
        Qualitative results on the Occ3D-nuScenes validation split.
        Owing to its persistent temporal representation, SGE more effectively preserves static structures under occlusion.
        As the side street branching to the upper left becomes occluded, LaGS loses the extent of the road and the traffic island, whereas SGE retains most of the geometry.
        Only the most informative camera view is shown.
    }%
    \label{fig:qualitative_nusc}
    \vspace{-0.4cm}
\end{figure}

\begin{figure}[tbh]
    \centering
    \input{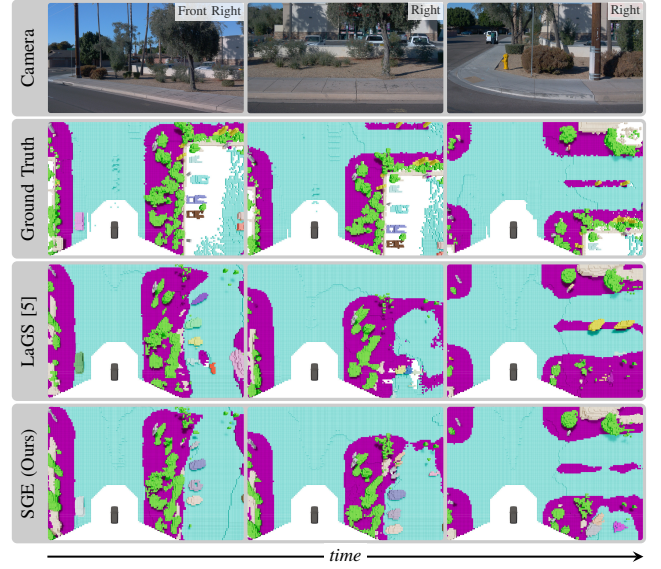}%
    \vspace{-0.75em}
    \caption{
        Qualitative results on the Occ3D-Waymo validation split.
        As occlusion from roadside vegetation increases, SGE preserves both parked vehicles and the surrounding parking lot structure.
        In contrast, LaGS gradually degrades object geometry and ultimately collapses the parking area.
        Only the most informative camera view is shown.
    }%
    \label{fig:qualitative_waymo}
\end{figure}

Qualitative comparisons on the Occ3D-nuScenes and Occ3D-Waymo datasets are shown in \cref{fig:qualitative_nusc,fig:qualitative_waymo}.
Notably, our streaming Gaussian encoder yields more stable volumetric predictions under partial observations and occlusions.
On nuScenes, the persistent representation enables the model to retain static scene layout even after regions become occluded by vegetation or terrain, whereas LaGS progressively loses road extent and fine structural details.
On Waymo, similar behavior is observed in cluttered parking scenarios, where our method better preserves both parked vehicles and surrounding structures under vegetation and structural occlusions.
These results indicate that enforcing temporal coherence at the representation level improves robustness to occlusion and promotes consistent scene modeling over time.

\subsection{Ablation Studies}
\label{ssec:ablations}

\begin{table}[t]
    \centering
    \caption{Ablation Study of Individual Components}
    \vspace{-0.75em}
    \footnotesize
\setlength{\tabcolsep}{2.5pt}  

\begin{tabular}{ccccc*{5}{+S[table-format=2.1,detect-all]}}
\toprule
\multicolumn{5}{c}{\textbf{Design Choices}} &
& & \multicolumn{2}{c}{mIoU} & \\
\cmidrule(lr){1-5} \cmidrule(lr){8-9}
EMC & DS & CL & Pruning & Sampling &
{\STQs} & {\AQs} & {Things} & {Stuff} & {IoU} \\

\midrule    
\xmark & \xmark & \xmark & $\mathcal{M}(r)$            & $\|\tensor{f}\|$   &        35.8  &        32.7  &        35.0  &        45.4  &        67.6  \\
\xmark & \cmark & \xmark & $\mathcal{M}(r)$            & $\|\tensor{f}\|$   &        35.9  &        32.8  &        34.9  &        45.4  &        67.6  \\
\cmark & \xmark & \xmark & $\mathcal{M}(r)$            & $\|\tensor{f}\|$   &        37.9  &        34.8  &        37.3  &        47.0  &        68.2  \\
\rowcolor{tabhl}
\cmark & \cmark & \xmark & $\mathcal{M}(r)$            & $\|\tensor{f}\|$   & \tabbf{38.4} & \tabbf{35.1} & \tabbf{37.9} & \tabul{47.9} & \tabbf{69.8} \\
\cmark & \cmark & \cmark & $\mathcal{M}(r)$            & $\|\tensor{f}\|$   & \tabul{38.3} & \tabul{34.9} &        37.6  & \tabbf{48.0} &        69.0  \\

\midrule    
\cmark & \cmark & \xmark & $\mathcal{M}(\alpha)$       & $\|\tensor{f}\|$   &        38.0  &        34.8  &        37.3  &        47.4  &        68.8  \\
\cmark & \cmark & \xmark & $\mathcal{M}(r_\text{acc})$ & $\|\tensor{f}\|$   &        37.8  &        34.7  &        37.1  &        47.0  &        69.2  \\
\cmark & \cmark & \xmark & $\operatorname{top}_k(r)$   & $\|\tensor{f}\|$   &        38.1  &        34.8  & \tabul{37.8} &        47.4  & \tabul{69.3} \\

\midrule    
\cmark & \cmark & \xmark & $\mathcal{M}(r)$            & $\|\tensor{f}\|{\cdot}e^{\shortminus 2 \tensor{d}}$    & \tabbf{38.4} & \tabbf{35.1} & \tabul{37.8} & 47.8 & \tabbf{69.8} \\
\cmark & \cmark & \xmark & $\mathcal{M}(r)$            & $\|\tensor{f}\|{\cdot}e^{\shortminus 5 \tensor{d}}$    &        38.0  &        34.5  &        37.6  & 47.8 & \tabul{69.3} \\

\bottomrule
\end{tabular}

\vspace{0.5em}
\begin{minipage}[t]{0.99\linewidth}
    \scriptsize
    Performance reported after single-frame pre-training (12 epochs) on Occ3D nuScenes.
    \textit{EMC:} Ego-motion compensation.
    \textit{DS:} Depth supervision.
    \textit{CL:} Consistency loss.
    For pruning: $\mathcal{M}(r)$ denotes multinomial sampling based on hybrid ranks, $\operatorname{top}_k(r)$ indicates deterministic rank-based retention, $r_{\text{acc}}$ replaces confidence with a learned accuracy score when computing the hybrid rank.
    For query initialization: $\|\tensor{f}\|$ samples proportionally to feature magnitude, $\|\tensor{f}\| e^{-2\tensor{d}}$ incorporates density-aware weighting.
\end{minipage}%

    \label{tab:ablations_components}
\end{table}

We conduct controlled ablations to quantify the contribution of each component of our proposed streaming Gaussian encoder.
Specifically, we analyze (i) core architectural elements, (ii) pruning strategies, (iii) query initialization schemes, and (iv) temporal context length.
Unless stated otherwise, results are reported on the Occ3D-nuScenes validation split.


\begin{figure}[tbh]
    \centering
    \input{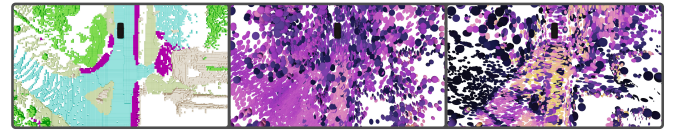}%
    \vspace{-0.5em}
    \caption{
        Gaussian opacity as a proxy for visibility.
        From left to right: ground-truth semantic occupancy, predicted opacities without depth supervision, and with depth supervision (darker indicates lower opacity).
        Depth supervision aligns opacity with geometric visibility, producing characteristic \enquote{shadows} in occluded regions (e.g., the road on the left).
    }%
    \label{fig:opacity}
    \vspace{-0.4cm}
\end{figure}

{\parskip=2pt
\noindent\textit{Core Architectural Components}: 
Component ablations are shown in \cref{tab:ablations_components} (top block).
Starting from a baseline without ego-motion compensation (EMC) or depth supervision (DS), adding EMC substantially improves \STQs and \AQs, highlighting the importance of spatial alignment for persistent state propagation.
Without EMC, propagated queries drift, limiting temporal aggregation.
Adding DS further improves semantic and geometric metrics by shaping opacity to reflect geometric visibility (cf.~\cref{fig:opacity}), strengthening visibility-driven confidence accumulation.
Together, EMC and DS produce clear cumulative gains.
We additionally evaluate an explicit temporal consistency loss (CL) on Gaussian parameters, enforcing cross-frame consistency of opacity, geometry, and semantic logits via $\mathcal{L}_2$ and KL penalties.
However, this provides no measurable benefit.
As the model is trained in a streaming regime with repeated query propagation and refinement, it already learns stable update dynamics.
Thus, representation-level temporal coherence emerges naturally from streaming refinement and visibility aggregation, rendering additional parameter-level constraints unnecessary.
}

{\parskip=2pt
\noindent\textit{Pruning Strategy}:
We next compare pruning variants (\cref{tab:ablations_components}, middle block).
Replacing hybrid rank-based pruning $\mathcal{M}(r)$ with opacity-only pruning $\mathcal{M}(\alpha)$ consistently reduces performance, suggesting that instantaneous visibility alone is insufficient for robust state management.
Substituting confidence with a learned semantic accuracy score, $\mathcal{M}(r_{\text{acc}})$, leads to a further performance drop, indicating that accumulated visibility provides a stronger signal for query retention than instantaneous correctness estimates.
Finally, deterministic retention $\operatorname{top}_k(r)$ performs slightly worse than stochastic multinomial sampling, suggesting that while the hybrid ranking criterion is the primary contributor to performance, stochasticity provides a small additional benefit.
}

\begin{table}[t]
    \centering
    \caption{Impact of Training Context Length and Consistency Loss}
    \vspace{-0.75em}
    \footnotesize
\setlength{\tabcolsep}{6pt}

\begin{tabular}{c*{6}{+S[table-format=2.1,detect-all]}}
\toprule
 & & & \multicolumn{3}{c}{mIoU} \\
\cmidrule(lr){4-6}
Warm-up & {\STQs} & {\AQs} & {All} & {Things} & {Stuff} & {IoU} \\
\midrule


2 &        36.1  &        32.3  &        39.5  &        35.1  &        45.9  &        68.3  \\
4 &        37.3  &        34.0  &        40.9  &        36.8  &        46.8  &        68.5  \\
6 &        38.0  & \tabul{34.7} & \tabul{41.6} &        37.3  &        47.6  & \tabul{69.5} \\
\rowcolor{tabhl}
7 & \tabbf{38.4} & \tabbf{35.1} & \tabbf{42.0} & \tabbf{37.9} & \tabul{47.9} & \tabbf{69.8} \\
8 & \tabul{38.1} & \tabul{34.7} & \tabbf{42.0} & \tabul{37.7} & \tabbf{48.0} & \tabbf{69.8} \\

\bottomrule
\end{tabular}

\vspace{0.5em}
\begin{minipage}[t]{0.88\linewidth}
    \scriptsize
    Performance reported after single-frame pre-training (12 epochs) on Occ3D nuScenes.
    \tinycolorbox{tabhl}{Shaded blue background} indicates our chosen configuration.
\end{minipage}%
\vspace{-0.5em}
    \label{tab:ablations_temporal}
\end{table}

{\parskip=2pt
\noindent\textit{Query Initialization}:
By default, new queries are initialized via multinomial sampling proportional to feature magnitude $\|\tensor{f}\|$, favoring well-supported regions.
To encourage exploration of newly observed or underexplored regions, we additionally evaluate density-aware weighting $\|\tensor{f}\| \cdot e^{-\beta \tensor{d}}$, where $\tensor{d}$ is the accumulated Gaussian density from the previous timestep.
This biases initialization towards uniform coverage (\cref{tab:ablations_components}, bottom block).
However, we find that moderate density-aware weighting performs nearly identically to magnitude-based initialization while stronger weighting degrades performance, suggesting that confidence-guided pruning and iterative refinement already provide sufficient adaptability.
}

{\parskip=2pt
\noindent\textit{Temporal Context Size}:
We analyze the number of warm-up frames used to initialize the persistent state (\cref{tab:ablations_temporal}).
Increasing the context from 2 to 7 frames consistently improves performance, after which gains largely saturate.
We therefore adopt 7 warm-up frames as a practical trade-off between performance and training efficiency.
}

\section{Conclusion}
\label{sec:conclusion}

We presented a streaming Gaussian encoder for camera-based 4D panoptic occupancy tracking that maintains a persistent and temporally coherent volumetric scene representation.
Moving beyond frame-centric feature recomputation, our method propagates and updates a fixed-budget set of latent Gaussian queries, enabling representation-level temporal consistency with negligible inference overhead.
Central to our approach is the interpretation of Gaussian opacity as a proxy for per-frame visibility, whose accumulation over time enables principled confidence-driven state management.
Together with ego-motion-aware propagation and streaming refinement, this results in a compact yet expressive scene representation that remains stable under occlusion, partial observation, and dynamic changes.
Extensive experiments on Occ3D-extended nuScenes and Waymo demonstrate state-of-the-art performance for camera-based 4D-POT.
More broadly, our findings highlight the potential of persistent volumetric representations as a step toward continuous 4D scene modeling in camera-only systems.

{
    \small
    \bibliographystyle{IEEETran}
    \bibliography{root}
}

\end{document}